\documentclass{article}

\usepackage{microtype}
\usepackage{graphicx}
\usepackage{subfigure}
\usepackage{booktabs} 

\usepackage{hyperref}

\usepackage{amsmath}
\usepackage{amssymb}
\usepackage{bbm}

\usepackage{xspace}

\newcommand{\RSet}{\mathbb{R}}

\newcommand{\lds}{\textsc{LDS}\xspace}

\usepackage[accepted]{icml2019}

\icmltitlerunning{Learning Discrete Structures for Graph Neural Networks}

\begin{document}

\twocolumn[
\icmltitle{Learning Discrete Structures for Graph Neural Networks}




\begin{icmlauthorlist}
\icmlauthor{Luca Franceschi}{iit,ucl}
\icmlauthor{Mathias Niepert}{nec}
\icmlauthor{Massimiliano Pontil}{iit,ucl}
\icmlauthor{Xiao He}{nec}
\end{icmlauthorlist}

\icmlaffiliation{nec}{NEC Labs Europe, Heidelberg, Germany. Work done in part as Luca Franceschi was visiting researcher at NEC}

\icmlaffiliation{iit}{CSML, Istituto Italiano di Tecnologia, Genoa, Italy.}

\icmlaffiliation{ucl}{University College London, London, UK.}

\icmlcorrespondingauthor{Luca Franceschi}{luca.franceschi@iit.it}
\icmlcorrespondingauthor{Xiao He}{xiao.he@neclab.eu}

\icmlkeywords{Machine Learning, ICML}

\vskip 0.3in
]



\printAffiliationsAndNotice{}

\begin{abstract}
Graph neural networks (GNNs) are a popular class of machine learning models whose major advantage is their ability to incorporate a sparse and discrete dependency structure between data points. Unfortunately, GNNs can only be used when such a graph-structure is available. In practice, however, real-world graphs are often noisy and incomplete or might not be available at all. 
With this work, we propose to jointly learn the graph structure and the parameters of graph convolutional networks (GCNs) by approximately solving a bilevel program that learns a discrete probability distribution on the edges of the graph.   
This allows one to apply GCNs not only in scenarios where the given graph is incomplete or corrupted but also in those where a graph is not available.
We conduct a series of experiments that analyze the behavior of the proposed method and demonstrate that it outperforms related methods by a significant margin.
Code is available at \url{https://github.com/lucfra/LDS-GNN}.
\end{abstract}

\vspace{-6mm}

\section{Introduction}

Relational learning is concerned with methods that cannot only leverage  the attributes of data points but also their relationships. Diagnosing a patient, for example, not only depends on the patient's vitals and demographic information but also on the same information about their relatives, the information about the hospitals they have visited, and so on. Relational learning, therefore, does not make the assumption of independence between data points but models their dependency explicitly.   
Graphs are a natural way to represent relational information and there is a large number of learning algorithms leveraging graph structure. 
Graph  neural  networks  (GNNs) \citep{scarselli2009graph} are  one such class of algorithms that are able to incorporate sparse and discrete dependency structures between data points. 

While a graph structure is available in some domains, in others it has to  be inferred or constructed.
A possible approach is to first create a $k$-nearest neighbor ($k$NN) graph based on some measure of similarity between data points. This is a common strategy used by several learning methods such as LLE \citep{Roweis:2000} and Isomap \citep{tenenbaum2000global}. A major shortcoming of this approach, however, is that the efficacy of the resulting models hinges on the choice of $k$ and, more importantly, on the choice of a suitable similarity measure over the input features. In any case, the graph creation and parameter learning steps are independent and require heuristics  and trial and error. Alternatively, one could simply use a kernel matrix to model 
the similarity of examples implicitly
at the cost of introducing 
a dense dependency structure. 





With this paper, we follow a different route with the aim of learning \emph{discrete} and \emph{sparse} dependencies between data points while simultaneously training the parameters of  graph convolutional networks (GCN), a class of GNNs. Intuitively, GCNs learn node representations by passing and aggregating messages between neighboring nodes~\citep{kipf2016semi,monti2017geometric,gilmer2017neural,hamilton2017inductive,duran2017learning,velickovic2017graph}. 
We propose to learn a generative probabilistic model for graphs, samples from which are used both during training and at prediction time. 
Edges are modelled with random variables whose parameters are treated as hyperparameters in a bilevel learning framework \citep{franceschi2018bilevel}. We iteratively sample the structure while minimizing an inner objective (a training error) 
and optimize the edge distribution parameters by minimizing an outer objective (a validation error).

To the best of our knowledge, this is the first method that simultaneously learns the graph and the parameters of a GNN for semi-supervised classification. Moreover, and this might be of independent interest, we adapt gradient-based hyperparameter optimization to work for a class of discrete hyperparameters (edges, in this work). 
We conduct a series of experiments and show that the proposed method is competitive with and often outperforms existing  approaches.  We also verify that the resulting graph generative models have meaningful edge probabilities. 

	




\section{Background}

We first provide some background on graph theory, graph neural networks, and bilevel programming.

\subsection{Graph Theory Basics}

A graph $G$ is a pair $(V, E)$ with $V = \{v_1, ..., v_N\}$ the set of vertices and $E \subseteq V \times V$ the set of edges. Let $N$ and $M$ be the number of vertices and edges, respectively. 
 Each graph can be represented by an adjacency matrix $A$ of size $N \times N$: $A_{i,j} = 1$ if there is an edge from vertex $v_i$ to vertex 
$v_j$, and $A_{i,j} = 0$ otherwise. 
The graph Laplacian is defined by $L = D - A$ where $D_{i,i} = \sum_{j} A_{i,j}$ and $D_{i,j} = 0$ if $i\neq j$. 
We denote the set of all $N\times N$ adjacency matrices by $\mathcal{H}_N$. 

\subsection{Graph Neural Networks} \label{sec:gnn:desc}

    Graph neural networks are a popular class of machine learning models for graph-structured data. We focus specifically on graph convolutional networks (GCNs) and their application to semi-supervised learning.
	All GNNs have the same two inputs. First,
	a feature matrix  $X\in \mathcal{X}_N\subset\mathbb{R}^{N\times n}$ where $n$ is the number of different node features, second, a graph $G=(V, E)$ with adjacency matrix $A \in \mathcal{H}_N$. 
	Given a set of class labels $\mathcal{Y}$ and a labeling function $
	y : V \rightarrow \mathcal{Y}$ 
	that maps (a subset of) the nodes to their true class label, the objective is, given a set of training nodes $V_{\mathtt{Train}}$, to learn a function
	$$f_{w}: \mathcal{X}_N \times \mathcal{H}_N \rightarrow \mathcal{Y}^{N}$$ by minimizing some regularized empirical loss 
	\begin{equation}  \label{eq:io_p1}
	L(w, A) = \sum_{v\in V_{\mathtt{Train}}} \ell(f_{w}(X, A)_v, y_v 
	) + \Omega(w),
	\end{equation}
	where $w\in\mathbb{R}^d$ are the parameters of $f_{w}$, $f_{w}(X, A)_v$ is the output of $f_{w}$ for node $v$, $\ell:\mathcal{Y}\times\mathcal{Y}\to\RSet^+$ is a point-wise loss function, and $\Omega$ is a regularizer.
	An example of the function $f_w$ proposed by \citet{kipf2016semi} is the following two hidden layer GCN that computes the class probabilities as 
\begin{equation}
\label{kipf-gcn}
    f_w(X, A) = \mathrm{Softmax}(\hat{A} \; \mathrm{ReLu}(\hat{A} \; X \; W_1) \; W_2),  
\end{equation}
where $w=(W_1, W_2)$ are the parameters of the GCN and $\hat{A}$ is the normalized adjacency matrix, given by  $\hat{A} = \tilde{D}^{-1/2} (A + I) \tilde{D}^{-1/2}$, with diagonal, $\tilde{D}_{ii} = 1 + \sum_j A_{ij}$.
	
	\subsection{Bilevel Programming in Machine Learning} \label{sec:bk:hpo}
	
Bilevel programs are optimization problems where a set of variables occurring in the objective function are constrained
to be an optimal 
solution of another optimization problem \citep[see][for an overwiew]{colson2007overview}. Formally
given two objective functions $F$ and $L$, the outer and inner objectives, and two sets of variables, $\theta\in\mathbb{R}^m$ and $w\in\mathbb{R}^d$, the outer and inner variables, a bilevel program is given by
\begin{equation}
\label{eq:bprog}
\min_{\theta, w_{\theta}} F(w_{\theta}, \theta) \ \ \text{such that} \ \  w_{\theta} \in \arg \min_w L(w, \theta).
\end{equation}
Bilevel programs arise in numerous situations such as  hyperparmeter optimization, adversarial, multi-task, and meta-learning~\citep{bennett2006model, flamary2014learning, munoz2017towards, franceschi2018bilevel}. 

Solving Problem \eqref{eq:bprog} is challenging since the solution sets of the inner problem are usually not available in closed-form. 
A standard approach
involves replacing the minimization of $L$ with the repeated application of an iterative optimization dynamics 
$\Phi$ such as (stochastic) gradient descent~ \citep{domke2012generic, maclaurin2015gradient, franceschi2017forward}. 
Let $w_{\theta, T}$ denote the inner variables after $T$ iterations of the dynamics $\Phi$, that is, $w_{\theta, T}=\Phi(w_{\theta, T-1}, \theta)=\Phi(\Phi(w_{\theta, T-2}, \theta), \theta)$, and so on.
Now, if $\theta$ and $w$ are real-valued and the objectives and dynamics smooth, we can compute the gradient of the function $F(w_{\theta, T}, \theta)$ w.r.t. $\theta$, 
denoted throughout as the \emph{hypergradient} $\nabla_{\theta} F(w_{\theta, T}, \theta)$, as
\begin{equation}
\label{eq:hypergrad1}
      \partial_{w} F(w_{\theta,T},\theta) 
    \nabla_{\theta} w_{\theta, T} + 
      \partial_{\theta} F(w_{\theta, T}, \theta),
\end{equation}
where the symbol $\partial$ denotes the partial derivative (the Jacobian) and $\nabla$ either the gradient (for scalar functions) or the total derivative. The first term 
can be computed efficiently in time $O(T(d + m))$  with reverse-mode algorithmic differentiation \citep{griewank2008evaluating} by unrolling the optimization dynamics, repeatedly substituting $w_{\Phi, t} = \Phi(w_{\theta, t-1}, \theta)$ and applying the chain rule.
This technique allows 
to optimize a number of hyperparameters several orders of magnitude greater than classic methods for hyperparameter optimization \citep{feurer2018hyperparameter}.

\begin{figure*}[t!]
    \centering
    \includegraphics[width=1.0\textwidth]{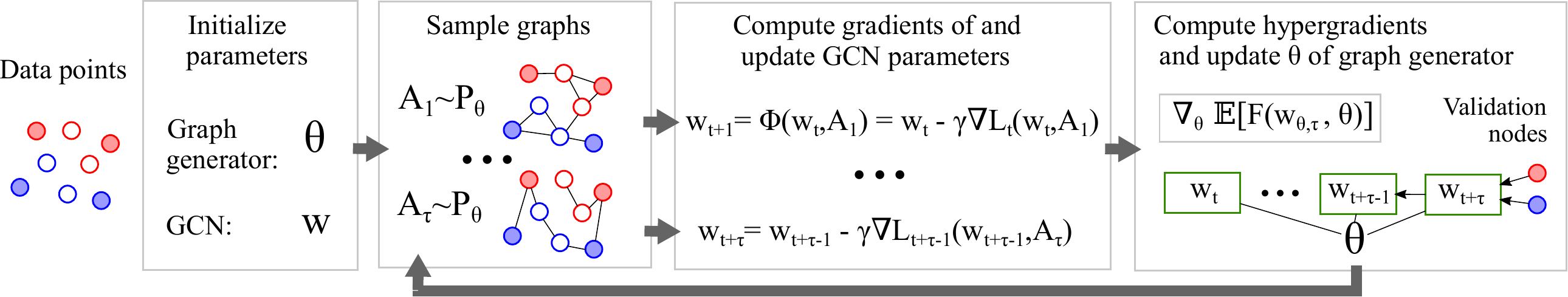}
    \vspace{-5mm}
    \caption{\label{fig:illustration}
    Schematic representation of our approach for learning discrete graph structures for GNNs.}
    \vspace{-5mm}
\end{figure*}

	\section{Learning Discrete Graph Structures} \label{sec:method}
	
    With this paper we address the challenging scenarios where a graph structure is either completely missing, incomplete, or noisy. 
   To this end, we learn a \emph{discrete} and \emph{sparse} dependency structure between data points while simultaneously training the parameters of a GCN. We frame this as a bilevel programming problem whose outer variables are the parameters of a generative probabilistic model for graphs. The proposed approach, therefore, optimizes both the parameters of a GCN and the parameters of a graph generator so as to minimize the classification error on a given dataset.     
We developed a practical algorithm based on truncated reverse-mode algorithmic differentiation \citep{williams1990efficient} and hypergradient estimation to approximately solve the bilevel problem. A schematic illustration of the resulting method is presented in Figure \ref{fig:illustration}. 

\subsection{Jointly Learning the Structure and Parameters}  \label{sec:lsst}

	Let us suppose that information about the true adjacency matrix $A$ is missing or incomplete. 
	Since, ultimately, we are interested in finding a model that minimizes the generalization error, we assume the existence of a second subset of instances with known target, $V_{\mathtt{Val}}$ (the validation set), from which we can estimate the generalization error. Hence, we propose to find
	$A\in\mathcal{H}_N$ that minimizes the function
    \begin{equation} \label{eq:oo_p1}
	F(w_A, A) = \sum_{v\in V_{\mathtt{Val}}} \ell(f_{w_A}(X, A)_v, y_v),
	\end{equation}
	where $w_A$ is the minimizer, assumed unique,
	of $L$ (see Eq. \eqref{eq:io_p1} and Sec. \ref{sec:bk:hpo}) for a \emph{fixed adjacency matrix} $A$.  
	We can then consider Equations \eqref{eq:io_p1} and \eqref{eq:oo_p1} as the inner and outer objective of a mixed-integer bilevel programming problem where the outer objective aims to find an optimal discrete graph structure and the inner objective the optimal parameters of a GCN given a graph. 
	
	The resulting bilevel problem is intractable to solve exactly even for small graphs. Moreover, it contains both continuous and discrete-valued variables,
which prevents us from directly applying Eq. \eqref{eq:hypergrad1}. A possible solution is to construct a continuous relaxation \citep[see e.g.][]{frecon2019bilevel}, another is to work with parameters of a probability distribution over graphs. The latter is the approach we follow in this paper.
	We maintain a generative model for the graph structure and reformulate the bilevel program in terms of the (continuous) parameters of the resulting distribution over discrete graphs. 
	Specifically, we propose to model each edge with a Bernoulli random variable.
Let $\overline{\mathcal{H}}_N = \mathrm{Conv}(\mathcal{H}_N)$ be the convex hull of the set of all adjacency matrices for $N$ nodes. 
	By modeling all possible edges as a set of mutually independent Bernoulli random variables with parameter matrix $\theta \in \overline{\mathcal{H}}_N$ we can sample graphs as $\mathcal{H}_N\ni A\sim \mathrm{Ber}(\theta)$.
	Eqs. \eqref{eq:io_p1} and \eqref{eq:oo_p1} can then be replaced, by using the expectation over graph structures. The resulting bilevel problem can be written as 
	\begin{eqnarray} 
		& \min_{\theta\in\overline{\mathcal{H}}_N} \mathbb{E}_{A\sim \mathrm{Ber}(\theta)}  \left[ F(w_{\theta}, A) \right] \label{eq:pp2_1}\\
		&\text{such that} \;\; w_{\theta} = \arg\min_w \mathbb{E}_{A\sim \mathrm{Ber}(\theta)}  \left[ L(w, A) \right]. \label{eq:pp2_2}
	\end{eqnarray}
	By taking the expectation, both the inner and the outer objectives become continuous (and possibly smooth) functions of the Bernoulli parameters. The bilevel problem given by Eqs. \eqref{eq:pp2_1}-\eqref{eq:pp2_2} is still challenging to solve efficiently. This is because the solution of the inner problem is not available in closed form for GCNs (the objective is non-convex); and the expectations are intractable to compute exactly\footnote{This is different than e.g. (model free) reinforcement learning, where the objective function is usually unknown, depending in an unknown way from the action and the environment.}.
	An efficient algorithm, therefore, will only be able to find approximate stochastic solutions, that is,  $\theta_{i,j}\in(0, 1)$. 

    Before describing a method to solve the optimization problem given by Eqs. \eqref{eq:pp2_1}-\eqref{eq:pp2_2} approximately with hypergradient descent, we first turn to the question of obtaining a final GCN model that we can use for prediction.
	For a given distribution $P_\theta$ over graphs with $N$ nodes and with parameters $\theta$, the expected output of a GCN is 
	\begin{equation}  \label{eq:model_exp}
 f^{\mathrm{exp}}_w(X) =  \mathbb{E}_{A
 } [f_w(X, A)]
	= \hspace{-2mm} \sum_{A \in \mathcal{H}_N} \hspace{-2mm} P_{\theta}(A) f_w(X, A).
	\end{equation}
	Unfortunately, computing this expectation is intractable even for small graphs; we can, however, compute an empirical estimate of the output as
	%
	%
	\begin{equation}
	\label{eq:empirical_mean_model}
	\hat{f}_w(X) = \frac{1}{S}\sum_{i=1}^S  f_w(X, A_i),
	\end{equation}
	where $S>0$ is the number of samples we wish to draw. Note that $\hat{f}$ is an unbiased estimator of $f^{\mathrm{exp}}_w$. Hence, to use a GCN $f_w$ learned with the bilevel formulation for prediction, we sample $S$ graphs from the distribution $P_\theta$ and compute the prediction as the empirical mean of the values of $f_w$. 
	
	Given the parametrization of the graph generator with Bernoulli variables ($P_{\theta} = \mathrm{Ber}(\theta)$), one can sample a new graph in $O(N^2)$. Sampling from a large number of Bernoulli variables, however, is highly efficient, trivially parallelizable, and possible at a rate of millions per second. Other sampling strategies such as MCMC sampling are possible in constant time.
	Given a set of sampled graphs, it is more efficient to evaluate a sparse GCN $S$ times than to use the Bernoulli parameters as weights of the GCN's adjacency matrix\footnote{Note also that $\mathbb{E} f_w(X, A) \neq f_w(X, \mathbb{E}  A) = f_w(X, \theta)$, as the model $f_w$ is, in general, nonlinear.
	}.
	Indeed, for GCN models, computing $\hat{f}_w$ has a cost of $O(SCd)$, rather than $O(N^2 d)$ for a fully connected graph, where $C=\sum_{ij} \theta_{ij}$ is the expected number of edges, and $d$ is the dimension of the weights. Another advantage of using a graph-generative model is that we can interpret it probabilistically which is not the case when learning a dense adjacency matrix.

	\subsection{Structure Learning via Hypergradient Descent}  \label{sec:hypergrad}
	
The bilevel programming 
formalism is a natural fit for the problem of learning both a graph generative model and the parameters of a GNN for a specific downstream task. Here, the outer variables $\theta$ are the parameters of the graph generative model and the inner variables $w$ are the parameters of the GCN. 

We now discuss a practical algorithm to approach the bilevel problem defined by Eqs.~\eqref{eq:pp2_1} and \eqref{eq:pp2_2}.
Regarding the inner problem, we note that the expectation 
\begin{equation} \label{eq:io_exp}
	 \mathbb{E}_{A\sim \mathrm{Ber}(\theta)}  \left[L(w, A) \right] = \sum_{A \in \mathcal{H}_N} P_{\theta}(A) L(w, A)
\end{equation}
is composed of a sum of $2^{N^2}$ terms, which is intractable even for relatively small graphs. We can, however, choose a tractable approximate learning dynamics $\Phi$ such as stochastic gradient descent (SGD),
\begin{equation} \label{eq:dyn:sgd}
w_{\theta, t+1} = \Phi(w_{\theta, t}, A_t)  =  w_{\theta, t} - \gamma_t \nabla L(w_{\theta, t}, A_t),
\end{equation}
where $\gamma_t$ is a learning rate and $A_t\sim\mathrm{Ber}(\theta)$ is drawn at each iteration.
Under appropriate assumptions and for $t\to\infty$, SGD  converges 
to a weight vector $w_{\theta}$ that depends on the edges' probability distribution \citep{bottou2010large}. 

Let $w_{\theta, T}$ be an approximate minimizer of $\mathbb{E} \left[ L \right]$ where $T$ may  depend on $\theta$.
We now need an estimator for the hypergradient $\nabla_{\theta} \mathbb{E}_{A\sim \mathrm{Ber}(\theta)} \left[ F(w_{\theta, T}, A) \right]$. 
Let us first consider the more general case of estimating $\nabla_{\theta} \mathbb{E}_{z\sim P_{\theta}} \left[h(z)\right]$ for some distribution $z\sim P_{\theta}$ with parameters $\theta$. If there exists a differentiable and reversible sampling path $\mathtt{sp}(\theta, \varepsilon)$ for  $P_{\theta}$, with $z= \mathtt{sp}(\theta, \varepsilon)$ for $\varepsilon\sim P_{\varepsilon}$, then one can use the general form of the pathwise gradient estimator~(see \citet{mohamed2019monte}, Sec. 5):
\begin{align}
\label{eq:pathwise:cont}
\nabla_{\theta} \mathbb{E}_{z\sim P_{\theta}} \left[h(z)\right] = 
\mathbb{E}_{\varepsilon \sim P_{\varepsilon}} \left[ \nabla_{\theta} h(\mathtt{sp}(\theta, \varepsilon)) \right] = \\
\mathbb{E}_{z\sim P_{\theta}} \left[ \nabla_z h(z) \nabla_{\theta} z\right].
\label{eq:pathwise:cont2}
\end{align}
Since we are concerned with discrete random variables, any sampling path would have discontinuities, making \eqref{eq:pathwise:cont}-\eqref{eq:pathwise:cont2} not directly applicable. 
Nevertheless, by using an inexact but smooth reparameterization for $P_{\theta}$, we may employ an approximate version of \eqref{eq:pathwise:cont}-\eqref{eq:pathwise:cont2} that allows us to derive a biased estimator 
 of the gradient $\nabla \mathbb{E}[h]$.
%
For $z = \mathtt{sp}(\theta, \varepsilon)=\theta$ the resulting gradient estimator is an instance of the class of 
straight-through estimators  (STE)~\cite{bengio2013estimating}.
%

Now, in our setting, we simply use the identity mapping 
$A=\mathtt{sp}(\theta, \varepsilon)=\theta$ 
%
%
and approximate 
\begin{multline}
\label{eq:inexact:pathwise}
\nabla_{\theta} \mathbb{E}_{A\sim \mathrm{Ber}(\theta)}\left[F(w_{\theta, T}, A)\right] \approx \\ 
\mathbb{E}_{A\sim \mathrm{Ber}(\theta)} \left[ \nabla_A F(w_{\theta, T}, A) \right].
\end{multline}
The second line instantiates \eqref{eq:pathwise:cont2} 
since
$\nabla_{\theta} A = \nabla_{\theta} \theta = \mathbf{I}$ with our choice of reparameterization for $P_{\theta}$.
This allows us to both take discrete samples in the forward pass and to use an efficient (low variance) pathwise gradient estimator in the reverse pass. The cost of this operation is the introduction of a bias, as 
setting $A=\mathtt{sp}(\theta, \varepsilon)=\theta$ is  not the same as sampling $A$ from $\mathrm{Ber}(\theta)$. 
Recalling equation \eqref{eq:hypergrad1}, we can further write $\mathbb{E}_{A\sim \mathrm{Ber}(\theta)} \left[ \nabla_A F(w_{\theta, T}, A) \right]$ as
\begin{equation} \label{eq:hypergrad:exp}
\mathbb{E}_{A
} \left[\partial_w F(w_{\theta, T}, A) \nabla_{A}
w_{\theta, T} + \partial_A F(w_{\theta, T}, A)\right]
\end{equation}
%
noting that $w_{\theta,T}$ depends on the distribution of $A$ through the optimization dynamics \eqref{eq:dyn:sgd}.
%
We then take the single sample Monte Carlo estimator of 
\eqref{eq:hypergrad:exp} to update the parameters $\theta$, projecting 
on the unit hypercube. 
We refer to this last quantity as the STE hypergradient, or simply hypergradient. 
We provide additional details about the computation of the term $\nabla_A w_{\theta, T}$ and the STE in the appendix.
%


Computing the STE hypergradient by fully unrolling the dynamics may be too expensive both in time and memory\footnote{Moreover, since we rely on biased estimations of the gradients, we do not expect to gain too much from a full computation.}. We propose to truncate the computation and estimate the hypergradient every $\tau$ iterations, where $\tau$ is a parameter of the algorithm. This is essentially an adaptation of truncated back-propagation through time \citep{werbos1990backpropagation, williams1990efficient} 
and can be seen as a short-horizon optimization procedure with warm restart on $w$.
%
%
%
A sketch of the method is presented in Algorithm \ref{alg:lsd}, while a more complete version that includes details on the STE hypergradient computation can be found in Appendix \ref{sec:apx:extalgo}. Inputs and operations in squared brackets are optional.

\begin{algorithm}[t]
\small
	\caption{\textbf{\lds}}
	\label{alg:lsd}
	\begin{algorithmic}[1]
		\STATE {\bfseries Input data:} $X$, $Y$, $Y'$[, $A$] 
		\STATE {\bfseries Input parameters:} $\eta$, $\tau$[, $k$] 
		\STATE $[A\gets\texttt{kNN}(X, k)]$ \hfill \COMMENT{Init. $A$ to $k$NN graph if $A=0$}
		\STATE $\theta \gets A$  \hfill \COMMENT{Initialize $P_\theta$ as a deterministic distribution} 
		\WHILE{Stopping condition is not met} 
			\STATE $t\gets 0$
			\WHILE{Inner objective decreases}
			\STATE$A_t \sim \mathrm{Ber}(\theta)$ 
\hfill \COMMENT{Sample structure} 
			\STATE $w_{\theta, t+1}\gets \Phi_t(w_{\theta, t}, A_t)$  \hfill \COMMENT{Optimize inner objective}
			\STATE $t\gets t + 1$
			\IF {$t = 0 \; (\mathrm{mod} \, \tau)$ {\bf or} $\tau=0$}
				\STATE $G \gets $ \texttt{computeHG}($F$, $Y$, $\theta$,  $(w_{\theta, i})_{i=t-\tau}^t$) 
				\STATE $\theta \gets \mathrm{Proj}_{\overline{\mathcal{H}}_N}[\theta - \eta G]$ \hfill \COMMENT{Optimize outer objective}
			\ENDIF
			\ENDWHILE
		\ENDWHILE
		\STATE {\bf return} $w$, $P_{\theta}$ \hfill \COMMENT{Best found weights and prob. distribution}
	\end{algorithmic}
\end{algorithm}

%
%
The algorithm contains stopping conditions at the outer and at the inner level. While it is natural to implement the latter with a decrease condition on the inner objective\footnote{We continue optimizing $L$ until $L(w_{t-1}, A) (1 + \varepsilon) \geq L(w_{\theta, t}, A)$, for $\varepsilon>0$ ($\varepsilon=10^{-3}$ in the experiments). Since $L$ is non-convex, we also use a patience window of $p$ steps.}, we find it useful to implement the first with a simple early stopping criterion. A fraction of the examples in the validation set is held-out to compute, in each outer iteration, the 
accuracy using the predictions of the empirically expected model~\eqref{eq:empirical_mean_model}. The optimization procedure terminates if there is no improvement for some consecutive outer loops. This helps avoiding overfitting the outer objective \eqref{eq:pp2_1}, which may be a concern in this context given the quantity of (hyper)parameters being optimized and the relative small size of the validation sets.

The STE hypergradients estimated with Algorithm \ref{alg:lsd} 
at each outer iteration are biased. The bias stems from both the straight-trough estimator and from the truncation procedure introduced in lines 11-13 \citep{tallec2017unbiasing}. Nevertheless, we find empirically that the algorithm is able to make reasonable progress, finding configurations in the distribution space that are beneficial for the tasks at hand. 
	

\begin{figure*}[t!]
    \centering
    \includegraphics[width=0.33\textwidth]{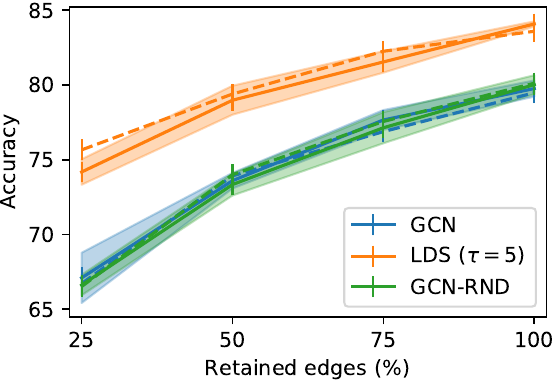}
    \includegraphics[width=0.33\textwidth]{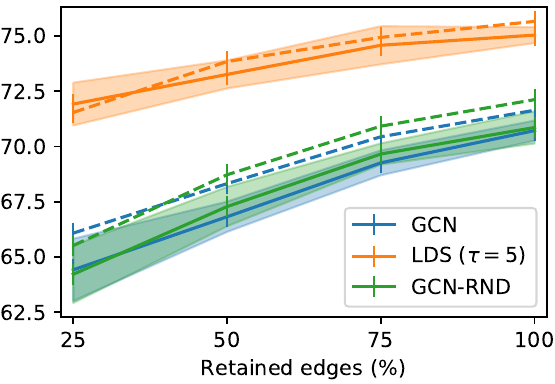}
    \includegraphics[width=0.33\textwidth]{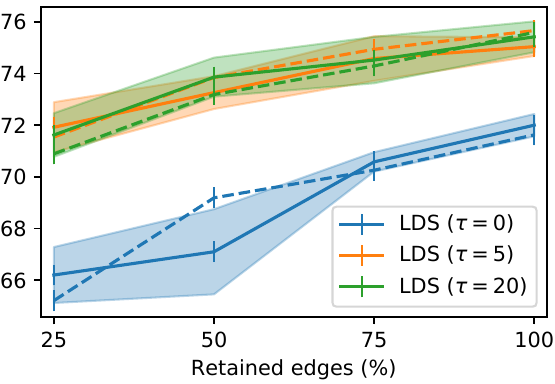}
    \vspace{-8mm}
    \caption{Mean accuracy $\pm$ standard deviation on validation (early stopping; dashed lines) and test (solid lines) sets for edge deletion scenarios on Cora (left) and Citeseer (center). 
    (Right) Validation of the number of
    steps $\tau$ 
    used to compute the STE hypergradient (Citeseer); $\tau=0$ corresponds to alternating minimization.
    All results are obtained from five runs with different random seeds.}
    \label{fig:edgrem}
    \vspace{-4mm}
\end{figure*}

\section{Experiments} \label{sec:experiments}


We conducted a series of experiments with three main objectives. First, we evaluated \lds on node classification problems where a graph structure is available but where a certain fraction of edges is missing.  
Here, we compared \lds with graph-based learning algorithms including vanilla GCNs.
Second, we wanted to validate our hypothesis that \lds can achieve competitive results on semi-supervised classification problems for which a graph is \emph{not} available. 
To this end, we compared \lds with a number of existing semi-supervised classification approaches. We also compared \lds with algorithms that first create $k$-NN affinity graphs on the data set.  Third, we analyzed the learned graph generative model to understand to what extent \lds is able to learn meaningful edge probability distributions even when a large fraction of edges is missing.

\subsection{Datasets}

Cora and Citeseer are two benchmark datasets that are commonly used to evaluate relational learners in general and GCNs in particular~\citep{sen2008collective}. 
The input features are bag of words and the task is node classification. 
We use the same dataset split and experimental setup of previous work \citep{yang2016revisiting, kipf2016semi}.
%
To evaluate the robustness of \lds on incomplete graphs, we construct graphs with missing edges by randomly sampling $25\%$, $50\%$, and $75\%$ of the edges. 
In addition to Cora and Citeseer where we removed all edges, we evaluate \lds on  benchmark datasets that are available in  scikit-learn~\citep{scikit-learn} such as Wine, Breast Cancer (Cancer), Digits, and 20 Newsgroup (20news). We take $10$ classes from 20 Newsgroup and use words (TFIDF) with a frequency of more than $5\%$ as features. 
We also use FMA, a dataset where $140$ audio features are extracted from 7,994 music tracks and where the problem is genre classification~\citep{fma_dataset}. The statistics of the datasets are reported in the appendix.

\subsection{Setup and Baselines}

For the experiments on graphs with missing edges, we compare \lds to vanilla GCNs. In addition, we also conceived a method (GCN-RND) where we add randomly sampled edges at each optimization step of a vanilla GCN. With this method we intend to show that simply adding random edges to the standard training procedure of a GCN model (perhaps acting as a regularization technique) is not enough to improve the generalization.
%

When a graph is completely missing, GCNs boil down to feed-forward neural networks. Therefore, we evaluate different strategies to induce a graph on both labeled and unlabeled samples by creating (1) a sparse Erd\H{o}s-R\'enyi random graph \citep{erdos1960evolution} (Sparse-GCN); (2) a dense graph with equal edge probabilities (Dense-GCN); (3) a dense RBF kernel on the input features (RBF-GCN); and (4) a sparse $k$-nearest neighbor graph on the input features ($k$NN-GCN).
For \lds we initialize the edge probabilities using the $k$-NN graph ($k$NN-\lds). 
We further include a dense version of \lds where we learn a dense similarity matrix 
($k$NN-\lds (dense)).
In this setting, we compare \lds to popular semi-supervised learning methods such as 
label propagation (LP) \citep{zhu2003semi}, manifold regularization (ManiReg) \citep{belkins206manifold}, and semi-supervised embedding (SemiEmb) \citep{weston2012deep}. 
ManiReg and SemiEmb are given a $k$-NN graph as input for the Laplacian regularization.
We also compare \lds to baselines that do not leverage a graph-structure such as logistic regression (LogReg), support vector machines (Linear and RBF SVM), random forests (RF), and feed-forward neural networks (FFNN). 
For comparison methods that need a $k$NN graph, $k \in \{2,3, \dots,20\}$ and the metric (Euclidean or Cosine) are tuned using validation accuracy. For $k$NN-LDS, $k$ is tuned from $10$ or $20$.

\begin{table*}[t]
\caption{Test accuracy ($\pm$ standard deviation) in percentage on various classification datasets. The best results and the statistical competitive ones (by paired t-test with $\alpha=0.05$) are in bold. All experiments have been repeated with 5 different random seeds. We compare $k$NN-\lds to several supervised baselines and semi-supervised learning methods. No graph is provided as input. $k$NN-\lds achieves high accuracy results on most of the datasets and yields the highest gains on datasets with underlying graphs (Citeseer, Cora).}
\label{tab:ssvsd:big}
\begin{center}
\begin{small}
\begin{tabular}{lccccccc}
\toprule
            & Wine          & Cancer & Digits        & Citeseer        & Cora            & 20news          & FMA             \\
\midrule
LogReg      & 92.1 (1.3) & \textbf{93.3 (0.5)}  & 85.5 (1.5) & 62.2 (0.0)       & 60.8 (0.0)       & 42.7 (1.7)   & 37.3 (0.7)   \\
Linear SVM  & 93.9 (1.6) & \textbf{90.6 (4.5)}  & 87.1 (1.8) & 58.3 (0.0)       & 58.9 (0.0)       & 40.3 (1.4)   & 35.7 (1.5)   \\
RBF SVM     & \textbf{94.1 (2.9)} & \textbf{91.7 (3.1)}  & 86.9 (3.2) & 60.2 (0.0)       & 59.7 (0.0)       & 41.0 (1.1)    & \textbf{38.3 (1.0)}   \\
RF          & 93.7 (1.6) & \textbf{92.1 (1.7)}  & 83.1 (2.6) & 60.7 (0.7)   & 58.7 (0.4)   & 40.0 (1.1)     & \textbf{37.9 (0.6)}   \\
FFNN    & 89.7 (1.9) & \textbf{92.9 (1.2)}  & 36.3 (10.3) & 56.7 (1.7)   & 56.1 (1.6)   & 38.6 (1.4)   & 33.2 (1.3)   \\
\midrule
LP   & 89.8 (3.7) & 76.6 (0.5)   & \textbf{91.9 (3.1)} & 23.2 (6.7)   & 37.8 (0.2)   & 35.3 (0.9)   & 14.1 (2.1)   \\
ManiReg & 90.5 (0.1) & 81.8 (0.1)  & 83.9 (0.1) & 67.7 (1.6) & 62.3 (0.9) & \textbf{46.6 (1.5)} & 34.2 (1.1) \\
SemiEmb & 91.9 (0.1) & 89.7 (0.1)  & \textbf{90.9 (0.1)} & 68.1 (0.1) & 63.1 (0.1) & \textbf{46.9 (0.1)} & 34.1 (1.9) \\
\midrule
Sparse-GCN   & 63.5 (6.6)  & 72.5 (2.9) & 13.4 (1.5)   & 33.1 (0.9)   & 30.6 (2.1)   & 24.7 (1.2) & 23.4 (1.4)  \\
Dense-GCN    & 90.6 (2.8) & 90.5 (2.7)  & 35.6 (21.8) & 58.4 (1.1)   & 59.1 (0.6)   & 40.1 (1.5)   & 34.5 (0.9)   \\
RBF-GCN    & 90.6 (2.3) & \textbf{92.6 (2.2)}  & 70.8 (5.5) & 58.1 (1.2)   & 57.1 (1.9)   &  39.3 (1.4)  & 33.7 (1.4)   \\
$k$NN-GCN    & 93.2 (3.1) & \textbf{93.8 (1.4)}  & \textbf{91.3 (0.5)} & 68.3 (1.3)   & 66.5 (0.4)   &  41.3 (0.6)  & \textbf{37.8 (0.9)}   \\
\midrule
$k$NN-\lds     & \textbf{97.3 (0.4)} & \textbf{94.4 (1.9)}  & \textbf{92.5 (0.7)} & \textbf{71.5 (1.1)}   & \textbf{71.5 (0.8)}   & \textbf{46.4 (1.6)}   & \textbf{39.7 (1.4)}  \\
\bottomrule
\end{tabular}
\end{small}
\end{center}
\vspace{-4mm}
\end{table*}

We use the two layers GCN given by Eq.~(\ref{kipf-gcn}) with $16$ hidden neurons and $\mathrm {ReLu}$ activation. Given a set of labelled training instances $V_{\mathtt{Train}}$ (nodes or examples) we use the regularized cross-entropy loss
\begin{equation}
    L(w, A) = - \hspace{-2mm} \sum_{v\in V_{\mathtt{Train}}} y_v \circ \log \left[ f_w(X, A)_{v} \right] + \rho ||w_1||^2,
\end{equation}
where $y_v$ is the one-hot encoded target vector for the $v$-th instance, $\circ$ denotes the element-wise multiplication and $\rho$ is a non-negative coefficient. As additional regularization technique we apply dropout \citep{srivastava2014dropout} with $\beta = 0.5$ as in previous work. We use Adam~\citep{kingma2014adam} for optimizing $L$, tuning the learning rate $\gamma$ from \{$0.005$, $0.01$, $0.02$\}. 
The same number of hidden neurons and the same activation is used for SemiEmb and FFNN. 

For \lds, we set the initial edge parameters $\theta_{i,j}$ to $0$ except for the known  edges (or those found by $k$NN) which we set to $1$. We then let all the parameters (including those initially set to $1$) to be optimized by the algorithm.
We further split the validation set evenly to form the validation (A) and early stopping (B) sets. As outer objective we use the un-regularized cross-entropy loss on (A) and  optimize it with stochastic gradient descent.
with exponentially decreasing learning rate. Initial experiments showed that accelerated optimization methods such as Adam or SGD with momentum underperform in this setting. We tune the step size $\eta$ of the outer optimization loop and the number of updates $\tau$ used to compute the truncated hypergradient. Finally, we draw $S=16$ samples to compute the output predictions (see Eq. \eqref{eq:empirical_mean_model}). For \lds and GCN, we apply early stopping with a window size of $20$ steps.

\lds was implemented in TensorFlow \citep{tensorflow2015-whitepaper} and is available at \url{https://github.com/lucfra/LDS}. The implementations of  the supervised baselines and LP are those from the scikit-learn python package \citep{scikit-learn}. GCN, ManiReg, and SemiEmb are implemented in Tensorflow. The hyperparameters for all the methods are selected through the validation accuracy.


\subsection{Results}
\label{sec:withgraph}


\begin{table}[t]
\small
\vspace{-2mm}
\begin{center}
\caption{Initial number of edges and expected number of sampled edges of learned graph by \lds.}
\vspace{1mm}
\label{tab:ense}
\begin{tabular}{lccccc}
\toprule
\% Edges    & 25\%   & 50\%   & 75\%   & 100\% \\
\midrule
Cora Initial     & 1357 & 2714 & 4071 & 5429 \\
Cora Learned     & 3635.6 & 4513.9 & 5476.9 & 6276.4 \\
\midrule
Citeseer Initial & 1183 & 2366 & 3549 & 4732\\
Citeseer Learned & 3457.4 & 4474.2 & 7842.5 & 6745.2 \\
\bottomrule
\end{tabular}
\end{center}
\vspace{-7mm}
\end{table}

\begin{figure*}[tb]
    \centering
    \includegraphics[width=0.33\textwidth]{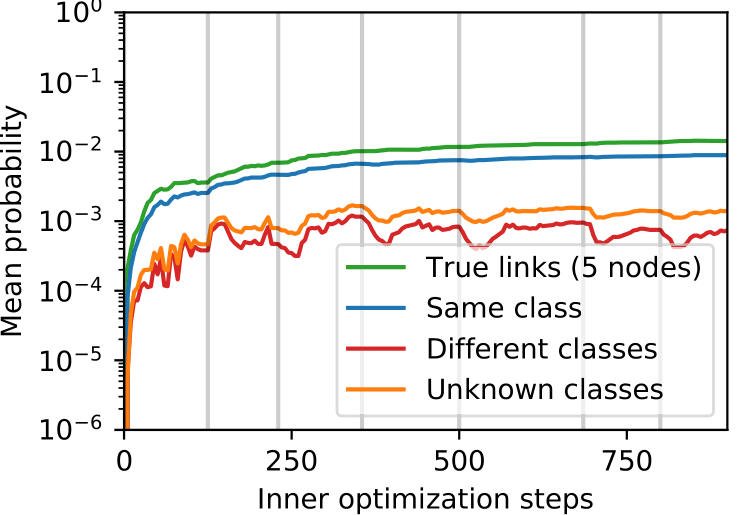}
    \includegraphics[width=0.33\textwidth]{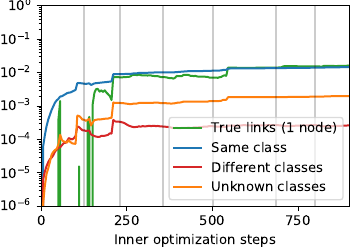}
    \includegraphics[width=0.33\textwidth]{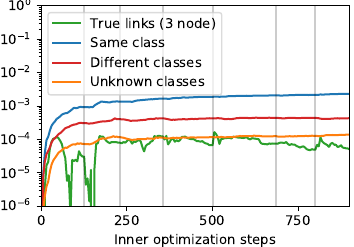}
    \vspace{-8mm}
    \caption{Mean edge probabilities to nodes aggregated w.r.t. four groups during \lds optimization, in $log_{10}$ scale for three example nodes. For each example node, all other nodes are grouped by the following criteria: (a) adjacent in the ground truth graph; (b) same class membership; (c) different class membership; and (d) unknown class membership. Probabilities are computed with \lds ($\tau=5$) on Cora with $25\%$ retained edges. From left to right, the example nodes belong to the training, validation, and test set, respectively. The vertical gray lines indicate when the inner optimization dynamics restarts, that is, when the weights of the GCN are reinitialized.}
    \label{fig:meanprobs}
    \vspace{-5mm}
\end{figure*}

The results on the incomplete graphs are shown in Figure \ref{fig:edgrem}  for Cora (left) and Citeseer (center). For each percentage of retained edges the accuracy on validation (used for early stopping) and test sets are plotted. 
\lds achieves competitive results in all scenarios and accuracy gains of up to $7$ percentage points. 
Notably, \lds improves the generalization accuracy of 
GCN models also when the given graph is that of the respective dataset (100\% of edges retained), by learning additional helpful edges. The accuracy of 84.1\% and 75.0\% for Cora and Citeseer, respectively, exceed all previous state-of-the-art results. Conversely, adding random edges does not help decreasing the generalization error. GCN and GCN-RND perform similarly which indicates that adding random edges to the graph is not helpful. 

Figure \ref{fig:edgrem} (right) depicts the impact of the number of iterations $\tau$ to compute the STE hypergradients. Taking multiple steps strongly outperforms alternating optimization\footnote{For $\tau=0$, one step of optimization of $L$ w.r.t. $w$, fixing $\theta$
is interleaved with one step of minimization of $F$ w.r.t. $\theta$, fixing $w$. 
Even if computationally lighter, this approach disregards the nested structure of \eqref{eq:pp2_1}-\eqref{eq:pp2_2}, not computing the first term of Eq. \eqref{eq:hypergrad1}.}
(i.e. $\tau=0$) in all settings. 
Increasing $\tau$ further to the value of $20$, however, does not yield significant benefits, while increasing the computational cost.

In Table \ref{tab:ense} we computed the expected number of edges in a sampled graph for Cora and Citeseer, to analyze the properties of the graphs sampled from the learned graph generator. The expected number of edges for \lds is higher than the original number which is to be expected since \lds has better accuracy results than the vanilla GCN in Figure \ref{fig:edgrem}. 
Nevertheless, the learned graphs are still very  sparse (e.g. for Cora, on average, less than $0.2\%$ edges are present). This facilitates efficient learning of the GCN 
in the inner learning loop of \lds. 

\begin{figure}[t!]
    \begin{center}
    \includegraphics[width=0.98\columnwidth]{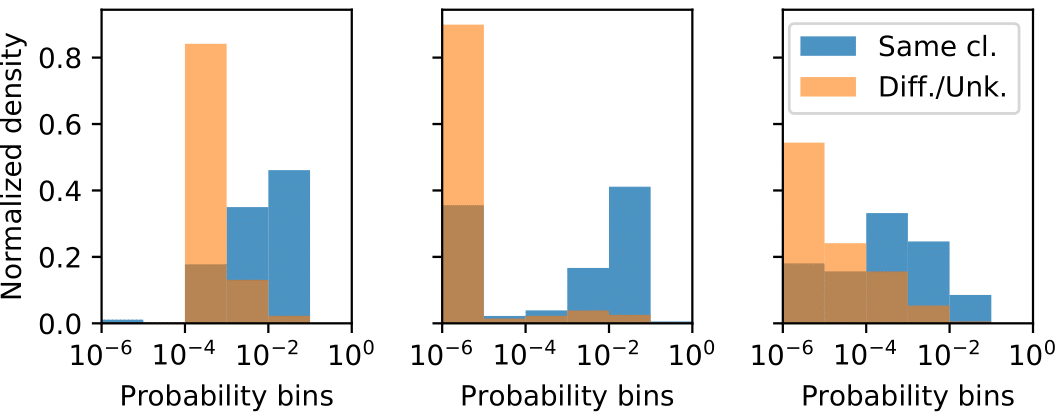}
    \end{center}
    \vspace{-4mm}
    \caption{Normalized histograms of edges' probabilities for the same nodes of Figure \ref{fig:meanprobs}.
    }
    \label{fig:histo}
    \vspace{-6mm}
\end{figure}

Table \ref{tab:ssvsd:big} lists the results for semi-supervised classification problems. The supervised learning baselines work well on some datasets such as Wine and Cancer but fail to provide competitive results on others such as Digits, Citeseer, Cora, and 20News. 
The semi-supervised learning baselines LP, ManiReg and SemiEmb can only improve the supervised learning baselines on $1$, $3$ and $4$ datasets, respectively. The results for the GCN with different input graphs show that $k$NN-GCN works well and provides competitive results compared to the supervised baselines on all datasets. 
$k$NN-\lds significantly outperforms $k$NN-GCN on $4$ out of the $7$ datasets. In addition, $k$NN-\lds is among the most competitive methods on all datasets and yields the highest gains on datasets that have an underlying graph. Moreover, $k$NN-\lds performs slightly better than its dense counterpart where we learn a dense adjacency matrix. The added benefit of the sparse graph representation lies in the potential to scale to larger datasets.

In Figure \ref{fig:meanprobs}, we show the evolution of mean edge probabilities during optimization on three types of nodes (train, validation, test) on the Cora dataset.  \lds is able to learn a graph generative model that is, on average, attributing $10$ to $100$ times more probability to edges between samples sharing the same class label. 
\lds often attributes a higher probability to edges that are present in the true held-out adjacency matrix (green lines in the plots).
In Figure \ref{fig:histo} we report the normalized histograms of the optimized edges probabilities for the same nodes of Figure \ref{fig:meanprobs}, sorted into six bins in $\log_{10}$-scale. Edges are divided in two groups: edges between nodes of the same class (blue) and between nodes of unknown or different classes (orange). \lds is able to learn highly non-uniform edge probabilities that reflect the class membership of the nodes. 
\begin{figure}[t!]
    \centering
    \includegraphics[width=0.98\columnwidth]{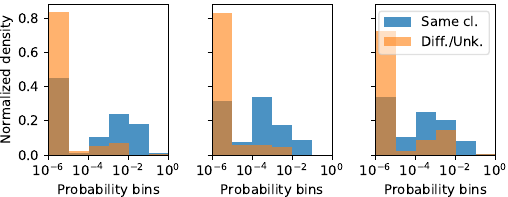}
    \vspace{-2mm}
    \caption{
    Histograms for three Citeseer test nodes, missclassified by $k$NN-GCN and rightly classified by $k$NN-\lds.}
    \label{fig:missnodesHist}
    \vspace{-6mm}
\end{figure}
Figure \ref{fig:missnodesHist} shows similar qualitative results as Figure \ref{fig:histo}, this time for three Citeseer test nodes, missclassified by $k$NN-GCN and correctly classified by $k$NN-\lds. Again, the learned edge probabilities linking to nodes of the same classes is significantly different to those from different classes; but in this case the densities are more skewed toward the first bin. 
On the datasets we considered, what seems to matter is to capture a useful distribution (i.e. higher probability for links between same class) rather than pick exact links; of course for other datasets this may vary. 

\section{Related work}

	
\textbf{Semi-supervised learning.} 
Early works on graph-based semi-supervised learning use graph Laplacian regularization
and include label propagation (LP) \citep{zhu2003semi}, manifold regularization (ManiReg) \citep{belkins206manifold}, and semi-supervised embedding (SemiEmb) \citep{weston2012deep}. 
These methods assume a given graph 
whose edges represent some similarity between nodes. Later, \citep{yang2016revisiting} proposed a method that uses graphs not for regularization but rather for embedding learning by jointly classification and graph context prediction. \citet{kipf2016semi} presented the first GCN for semi-supervised learning. There are now numerous GCN variants all of which assume a given graph structure. Contrary to all existing graph-based semi-supervised learning approaches, \lds{}
is able to work even when the graph is incomplete or missing.

\textbf{Graph synthesis and generation.} \lds learns a probabilistic generative model for graphs. The earliest probabilistic generative model for graphs was the Erd\H{o}s-R\'enyi random graph model \citep{erdos1960evolution}, where edge probabilities are modelled as identically distributed and mutually independent Bernoullis. Several  network models have been proposed to model well particular graph  properties such as degree distribution \citep{leskovec2005graphs} or network diameter \citep{watts1998collective}. 
\citet{leskovec2010kronecker} proposed a generative model based on the Kronecker product that takes a real graph as input and generates graphs that have similar properties.
Recently, deep learning based approaches have been proposed for graph generation \citep{you2018graphrnn,li2018learning,grover2018graphite,de2018molgan}. The goal of these methods, however, is to learn a sophisticated generative model that reflects the properties of the training graphs. \lds, on the other hand, learns graph generative models as a means to perform well on classification problems and its input is not a collection of graphs. More recent work proposed an unsupervised model that learns to infer interactions between entities while simultaneously learning the dynamics of physical systems such as spring systems~\citep{kipf2018neural}. 
Contrary to \lds, the method is specific to dynamical interacting systems, is unsupervised, and uses a variational encoder-decoder. 
Finally, we note that \citet{johnson2017learning} proposed a fully differentiable neural model able to process and produce graph structures at both input, representation and output levels; training the model requires, however, supervision in terms of ground truth graphs.  

\textbf{Link prediction.} Link prediction is a decades-old problem~\citep{liben2007link}. Several survey papers cover the large body of work ranging from link prediction in social networks to knowledge base completion~\citep{lu2011link,nickel2016review}. While a majority of the methods are based on some similarity measure between node pairs, there has been a number of neural network based methods \citep{zhang2017weisfeiler,zhang2018link}. The problem we study in this paper is related to link prediction as we also want to learn or extend a graph. 
However, existing link prediction methods do not simultaneously learn a GNN node classifier.
Statistical relational learning (SRL)~\citep{getoor2007introduction} models often perform both link prediction and node classification through the existence of binary and unary predicates. However, SRL models are inherently intractable and the structure  and parameter learning steps are  independent.

\textbf{Gradient estimation for discrete random variables.} Due to the intractable nature of the two bilevel objectives, \lds needs to estimate the hypergradients through a stochastic computational graph \citep{schulman2015gradient}. 
Using the score function estimator, also known as REINFORCE \citep{williams1992simple}, would treat the outer objective as a black-box function and would not exploit $F$ being differentiable w.r.t. the sampled adjacency matrices and inner optimization dynamics. Conversely, the path-wise estimator is not readily applicable, since the random variables are discrete. \lds borrows from a solution proposed before \citep{bengio2013estimating}, at the cost of having biased estimates. Recently, \citet{jang2016categorical, maddison2016concrete} presented an approach based on  continuous relaxations to reduce variance, which    \citet{tucker2017rebar} combined with REINFORCE to obtain an unbiased estimator.  \citet{grathwohl2017backpropagation} further introduced surrogate models to construct control variates for black-box functions. Unfortunately, these latter methods require to compute the function in the interior of the hypercube, possibly in multiple points \citep{tucker2017rebar}. This would introduce additional computational overhead\footnote{Recall that $F$ can be computed only after (approximately) solving the inner optimization problem.}. 




\section{Conclusion}

We propose \lds, a framework that simultaneously learns the graph structure and the parameters of a GNN. While we have used a specific GCN variant~\citep{kipf2016semi} in the experiments, the method is more generally applicable to other GNNs. 
The strengths of \lds are its high accuracy gains on typical semi-supervised classification datasets at a reasonable computational cost. 
Moreover, due to the graph generative model \lds learns, the edge parameters have a probabilistic interpretation. 

The method has its limitations. While relatively efficient, it cannot currently scale to large datasets: this would require an implementation that works with mini-batches of nodes.  We evaluated \lds only in the transductive setting,  when all data points (nodes) are available during training. Adding additional nodes after training (the inductive setting) would currently require retraining the entire model from scratch. When sampling graphs, we do not currently enforce the graphs to be connected. This is something we anticipate to improve the results, but this would require a more sophisticated sampling strategy. All of these shortcomings motivate future work.
In addition, we hope that suitable variants of \lds algorithm will also be applied to other problems such as neural architecture search or to tune other discrete hyperparameters.

\bibliography{bgl}
\bibliographystyle{icml2019}


\clearpage

\appendix

\section{Extended algorithm}
\label{sec:apx:extalgo}

In this section we provide an extended version of the Algorithm \ref{alg:lsd} that includes the explicit computation of the STE hypergradient by truncated reverse mode algorithmic differentiation. Recall that the inner objective is replaced by an iterative dynamics $\Phi$ such as stochastic gradient descent. Hence, starting from an initial point $w_0$, the iterates are computed as  
$$
w_{\theta, t+1} = \Phi(w_{\theta, t}, A_t).
$$
Let $D_t$ and $E_t$ denote the Jacobians of the dynamics:
$$
D_t := \partial_w \Phi(w_{\theta, t}, A_t); \quad E_t := \partial_A \Phi(w_{\theta, t}, A_t),
$$
and recall that, because of our reparameterization choice (for the backward pass), we have 
$
\nabla_{\theta} A = \nabla_{\theta} \theta = \mathbf{I}.
$
We report the pseudocode in Algorithm \ref{alg:lsd:ext}, where the letter $p$ is used to indicate the adjoint variables (Lagrangian multipliers).
Note that for $\tau=0$ the algorithm does not enter in the loop at line 15.
Finally, note also that at line 16, we re-sample the adjacency matrices instead of reusing  those computed in the forward pass (lines 8-10).

Algorithm \ref{alg:lsd:ext} was implemented in TensorFlow as an extension of the software package \texttt{Far-HO}, freely available at \url{https://github.com/lucfra/FAR-HO}.

\begin{algorithm}[t]
\small
	\caption{\textbf{\lds} (extended)}
	\label{alg:lsd:ext}
	\begin{algorithmic}[1]
		\STATE {\bfseries Input data:} $X$, $Y$, $Y'$[, $A$] 
		\STATE {\bfseries Input parameters:} $\eta$, $\tau$[, $k$] 
		\STATE $[A\gets\texttt{kNN}(X, k)]$ \hfill \COMMENT{Init. $A$ to $k$NN graph if $A=0$}
		\STATE $\theta \gets A$  \hfill \COMMENT{Initialize $P_\theta$ as a deterministic distribution} 
		\WHILE{Stopping condition is not met} 
			\STATE $t\gets 0$
			\WHILE{Inner objective decreases}
			\STATE$A_t \sim \mathrm{Ber}(\theta)$ 
\hfill \COMMENT{Sample structure} 
			\STATE $w_{\theta, t+1}\gets \Phi_t(w_{\theta, t}, A_t)$  \hfill \COMMENT{Optimize inner objective}
			\STATE $t\gets t + 1$
			\IF {$t = 0 \; (\mathrm{mod} \, \tau)$ {\bf or} $\tau=0$}
				\STATE$A_t \sim \mathrm{Ber}(\theta)$  
				\STATE $p\gets \partial_w F(w_{\theta, t}, A_t)$
				\STATE $G\gets \partial_A F(w_{\theta, t}, A_t)$
				\FOR{$s=t-1$ \textbf{downto} $t-\tau$}  
					\STATE $A_s \sim \mathrm{Ber}(\theta)$ 
				    \STATE $p\gets p D_s(w_{\theta, s}, A_s)$
				    \STATE $G \gets G + p E_s(w_{\theta, s}, A_s)$
				\ENDFOR 
				\STATE $\theta \gets \mathrm{Proj}_{\overline{\mathcal{H}}_N}[\theta - \eta G]$ \hfill \COMMENT{Optimize outer objective}
			\ENDIF
			\ENDWHILE
		\ENDWHILE
		\STATE {\bf return} $w$, $P_{\theta}$ \hfill \COMMENT{Best found weights and prob. distribution}
	\end{algorithmic}
\end{algorithm}




\section{On the Straight-through Estimator}
\label{sec:app:onSTE}

\lds{} borrows from an heuristic solution proposed before \citep{bengio2013estimating}, at the cost of having biased (hyper)gradient estimates. 
Given a function $h(z)$, where $z\sim P_{\theta}$ is a discrete random variable whose distribution depends on parameters $\theta$, the STE is a technique that consists in computing a biased estimator of the gradient of $\ell(\theta)=\mathbb{E}_{z\sim P_{\theta}} h(z)$ 
by using an inexact, but smooth, reparameterization for $z$, together with the application of an approximate version of \eqref{eq:pathwise:cont}-\eqref{eq:pathwise:cont2}. 

When $z$ is Bernoulli distributed, such reparameterization can be simply chosen\footnote{An exact, 
but discontinuous reparameterization for $z\sim\mathrm{Ber}(\theta)$ is, for instance, $z= \mathtt{sp}(\theta, \varepsilon)=H(\theta-\varepsilon)$ for $\varepsilon\sim\mathcal{U}(0, 1)$, where $H$ is the Heaviside function and $\mathcal{U}$ is the (continuous) uniform distribution.} 
as $z = \mathtt{sp}(\theta, \varepsilon)=\theta$  in which case the STE boils down to 
\begin{equation} \label{eq:stest}
    \hat{g}(z)=\frac{\partial h(z)}{\partial z}, \quad z\sim P_{\theta}.
\end{equation}
If $h$ is \emph{a smooth function of $z$}, Eq. \ref{eq:stest} is well defined and yields, in general, non-zero quantities. This operation may be viewed under different angles: e.g. as ``setting" $\frac{\partial z}{\partial \theta}$  to the identity, or, as ``ignoring" the hard thresholds in the backward pass. $\hat{g}$ is a random variable that depends, again, from $\theta$. The true gradient $\nabla \ell(\theta)$ can be estimated by drawing one or more samples from $\hat{g}$. 

As an illustrative example, consider the very simple case where $h(z)=(a z - b)^2/2$ for scalars $a$ and $b$, with $z\sim \mathrm{Ber}(\theta)$, $\theta\in[0,1]$. The gradient (derivative) of $\mathbb{E} \, \left[h \right]$ w.r.t. $\theta$ can be easily computed as 
\begin{multline*}
\frac{\partial}{\partial \theta} \, \mathbb{E}_{z\sim \mathrm{Ber}(\theta)} \left[ h(z) \right] = 
\\= \frac{\partial}{\partial \theta}  \left[ \theta \frac{(a-b)^2}{2} + (1-\theta) \frac{(-b)^2}{2} \right] 
= \frac{a^2}{2} - ab, \label{eq:st:true}
\end{multline*}
whereas the corresponding straight-through estimator, which is a random variable, is given by
\begin{equation*}
\hat{g}(z) = \frac{\partial h(z)}{\partial z} = (a z - b) a, \quad z \sim \mathrm{Ber}(\theta). \label{eq:st:est}
\end{equation*}
One has, however, that 
$$
\mathbb{E}_{z\sim \mathrm{Ber}(\theta)} \left[ \hat{g}(z) \right] = \theta (a-b)a + (1-\theta)(-ab) = \theta a^2 - ab,
$$
resulting in $\hat{g}$ to be biased for $\theta \neq \frac{1}{2}$.

\section{Additional Tables}

Table \ref{tb:data} contains the list of datasets we used in the experimental section, with relevant statistics.

\begin{table}[h]
\small
	\centering
	\vspace{-2mm}
	\caption{\label{tb:data}Summary statistics of the datasets.}
	\begin{tabular}{lccccc}
		\toprule Name & Samples & Features & $|\mathcal{Y}|$ & Train/Valid/Test \\ \midrule
		Wine & 178 & 13 &  3 & 10 / 20 / 158 \\
		Cancer & 569 & 30 &  2 & 10 / 20 / 539 \\
		Digits & 1,797 & 64 &  10 & 50 / 100 / 1,647 \\
		Citeseer & 3,327 & 3,703 &  6 & 120 / 500 / 1,000 \\ 
		Cora & 2,708 & 1,433 &  7 & 140 / 500 / 1,000 \\
		20news & 9,607 & 236 &  10 & 100 / 200 / 9,307 \\
		FMA & 7,994 & 140 & 8 & 160 / 320 / 7,514 \\ \bottomrule
	\end{tabular}
	\vspace{-2mm}
\end{table}

We report in Table \ref{tb:edg_ret_res} the numerical results relative to the experiments with various percentages of edges retained on Citeseer and Cora datasets. We refer to Section \ref{sec:experiments} for a complete description.
\begin{table*}[t]
    \centering
    \caption{Test accuracy ($\pm$ standard deviation) in percentage for the experiments on Citeseer and Cora, with various percentages of edges retained (see Figure \ref{fig:edgrem}). The data split is the same used in  \citep{kipf2016semi}}
    \vspace{2mm}
    \begin{tabular}{lccccc}
\toprule
\% Edges    & 25\%   & 50\%   & 75\%   & 100\% (full graph) \\
\midrule
GCN (Citeseer) & 64.42 (1.4) & 66.82 (0.7) & 69.26 (0.6) & 70.72 (0.5) \\
GCN-RND (Citeseer) & 64.22 (1.3) & 67.28 (0.9) & 69.66 (0.5) & 70.86 (0.7) \\
LDS (Citeseer) & 71.92 (1.0) & 73.26 (0.6) & 74.58 (0.9) & \textbf{75.04 (0.4)} \\
\midrule
GCN (Cora) & 67.10 (1.7) & 73.60 (0.5) & 77.66 (0.7) & 79.76 (0.5) \\
GCN-RND (Cora) & 66.62 (0.7) & 73.34 (0.7) & 77.14 (1.0) & 80.02 (0.6) \\
LDS (Cora) & 74.18 (1.0) & 78.98 (0.6) & 81.54 (0.9) & \textbf{84.08 (0.4)} 
\\
\bottomrule
\end{tabular}
    \label{tb:edg_ret_res}
\end{table*}

\section{Visualization of Embeddings}

\begin{figure*}[t]
    \centering
    \includegraphics[width=.33\textwidth]{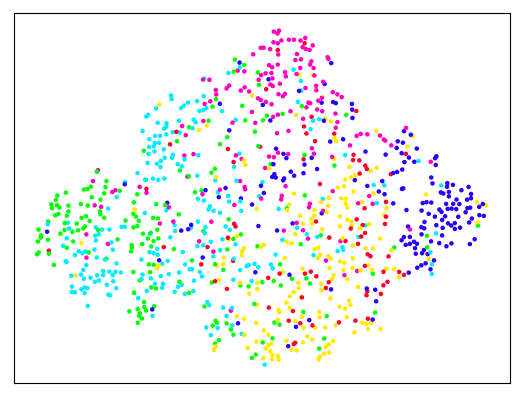}
    \hspace{-2mm}
    \includegraphics[width=.33\textwidth]{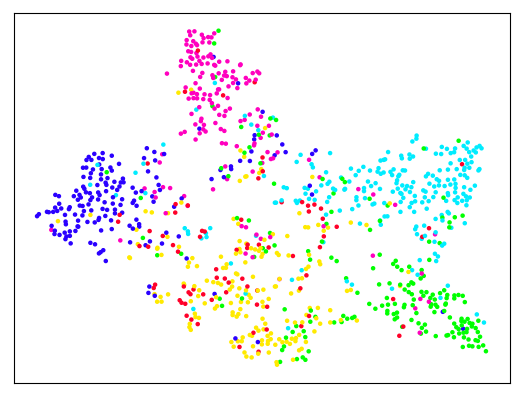}
    \hspace{-2mm}
    \includegraphics[width=.33\textwidth]{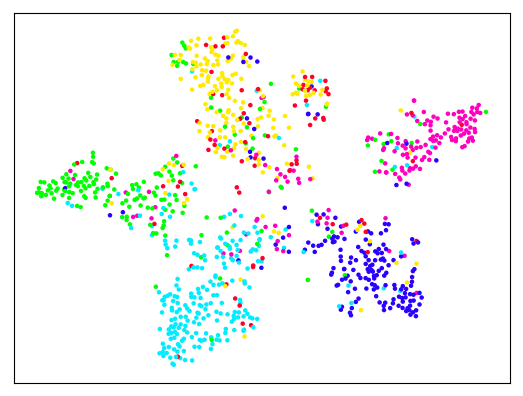}
    \caption{T-SNE visualization of the output activations (before the classification layer) on the Citeseer dataset. Left: Dense-GCN, Center: $k$NN-GCN, Right $k$NN-\lds}
    \label{fig:tsne}
\end{figure*}

We further visualize the embeddings learned by GCN and \lds using T-SNE \citep{maaten2008visualizing}. Figure \ref{fig:tsne} depicts the T-SNE visualizations of the embeddings learned on Citeseer with Dense-GCN (left), $k$NN-GCN (center), and $k$NN-\lds (right). As can be seen, the embeddings learned by $k$NN-\lds provides the best separation among different classes. 


%


\end{document}